\documentclass[10pt,twocolumn,letterpaper]{article}

\usepackage{iccv}
\usepackage{times}
\usepackage{epsfig}
\usepackage{graphicx}
\usepackage{amsmath}
\usepackage{amssymb}
\usepackage{threeparttable}
\usepackage{float}
\usepackage{multirow}
\usepackage{bbding}

\usepackage[pagebackref=true,breaklinks=true,letterpaper=true,colorlinks,bookmarks=false]{hyperref}

\iccvfinalcopy 

\def\iccvPaperID{9729} 

\begin{document}

\title{Divide-and-Assemble: Learning Block-wise Memory for Unsupervised Anomaly Detection}

\author{Jinlei Hou$^{1,2\dagger}$, Yingying Zhang$^1$, Qiaoyong Zhong$^1$, Di Xie$^1$, Shiliang Pu$^{1\star}$, Hong Zhou$^2$\\
$^1$Hikvision Research Institute\quad$^2$Zhejiang University\\
{\tt\small \{houjinlei,zhangyingying7,zhongqiaoyong,xiedi,pushiliang.hri\}@hikvision.com}\\
{\tt\small zhouh@mail.bme.zju.edu.cn}
}

\maketitle

\let\thefootnote\relax\footnotetext{$^\dagger$Work done as an intern at Hikvision Research Institute.}
\let\thefootnote\relax\footnotetext{$^\star$Corresponding author. This project was supported by the National Key R\&D Program of China (Grant No. 2020AAA010400X) and the National Natural Science Foundation of China (Grant No. 61803332).}

\begin{abstract}
  Reconstruction-based methods play an important role in unsupervised anomaly detection in images. Ideally, we expect a perfect reconstruction for normal samples and poor reconstruction for abnormal samples. Since the generalizability of deep neural networks is difficult to control, existing models such as autoencoder do not work well. In this work, we interpret the reconstruction of an image as a divide-and-assemble procedure. Surprisingly, by varying the granularity of division on feature maps, we are able to modulate the reconstruction capability of the model for both normal and abnormal samples. That is, finer granularity leads to better reconstruction, while coarser granularity leads to poorer reconstruction. With proper granularity, the gap between the reconstruction error of normal and abnormal samples can be maximized. The divide-and-assemble framework is implemented by embedding a novel multi-scale block-wise memory module into an autoencoder network. Besides, we introduce adversarial learning and explore the semantic latent representation of the discriminator, which improves the detection of subtle anomaly. We achieve state-of-the-art performance on the challenging MVTec AD dataset. Remarkably, we improve the vanilla autoencoder model by 10.1\% in terms of the AUROC score.
\end{abstract}

\section{Introduction}

\begin{figure}[t]
	\centering
	\includegraphics[width=\linewidth]{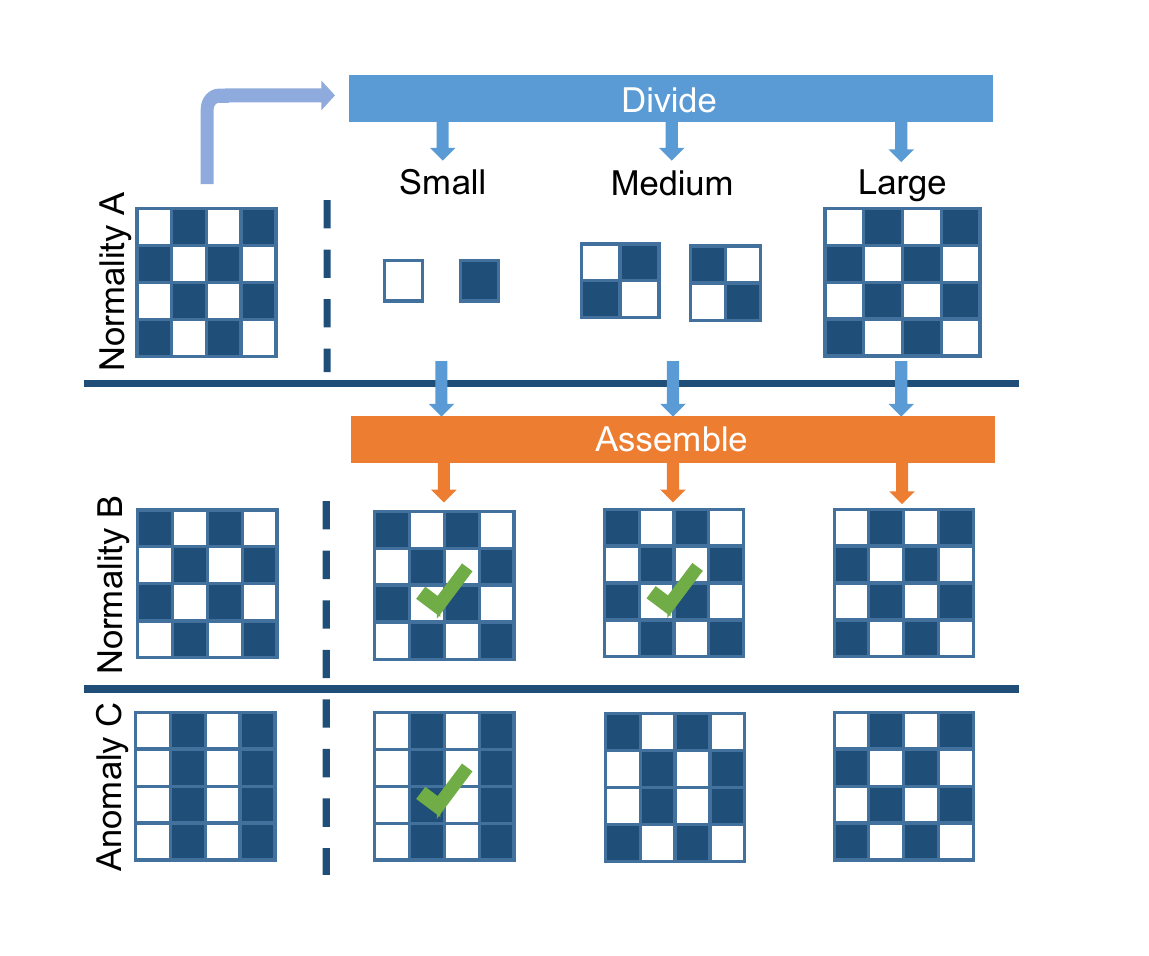}
	\caption{A toy example demonstrating that the reconstruction capability on normality and anomaly varies with different building block sizes. A: a normal sample with diagonal stripes. B: another normal sample with diagonal stripes similar to A. C: an abnormal sample with vertical stripes.}
	\label{toy-exp}
\end{figure}

Anomaly detection in images is an increasingly important area among the computer vision community. Lacking of abnormal training data makes this task challenging and limits the use of well-established supervised learning methods. Hence, existing works define anomaly detection as an unsupervised learning problem that tries to model the distribution of normality without abnormal samples at training time. During inference, samples described as outliers of the normality distribution are then considered to be anomalous.

\begin{figure}[t]
	\centering
	\includegraphics[width=\linewidth]{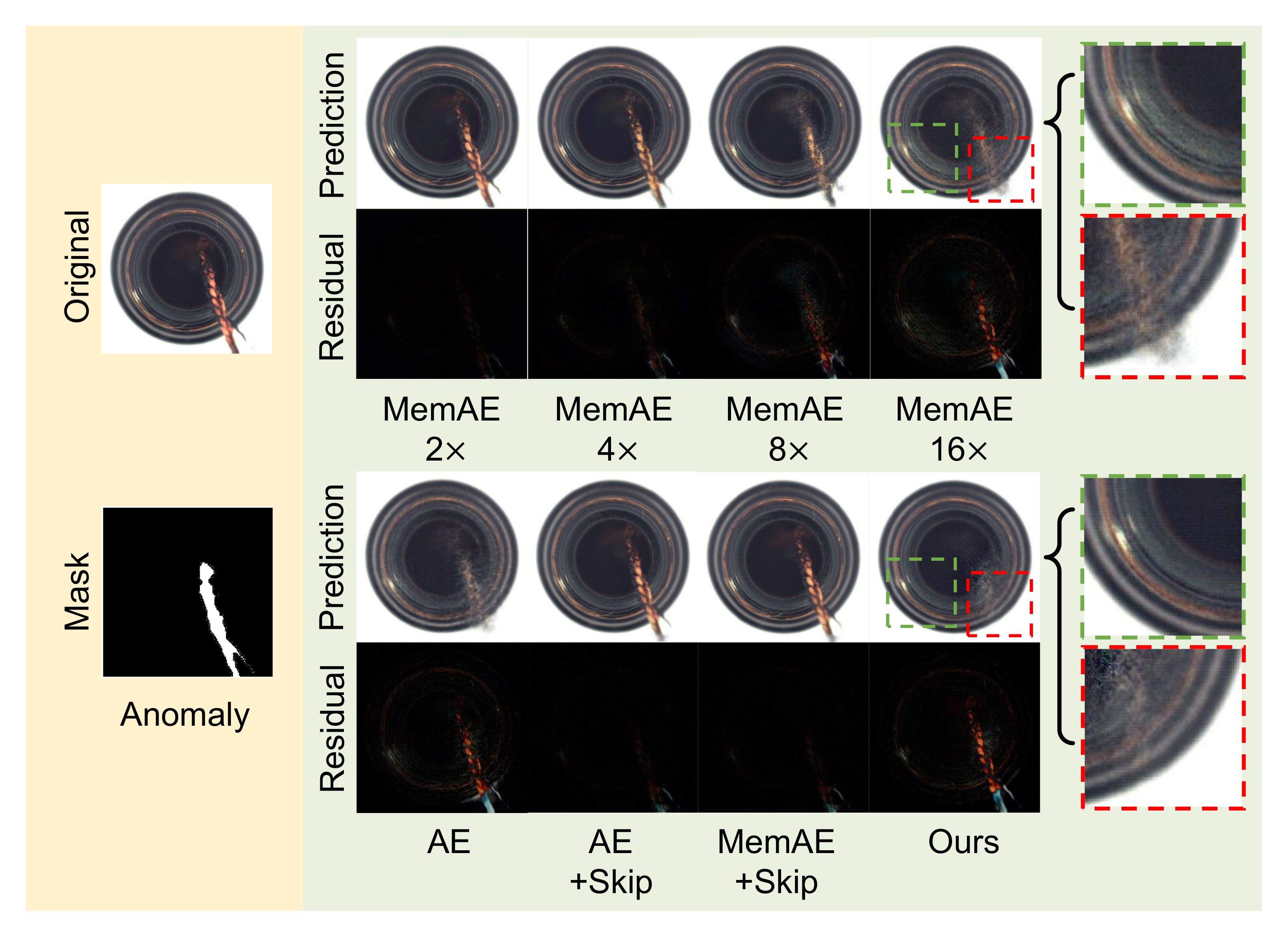}
	\caption{Exemplary reconstruction results of various models. \textbf{AE}: the vanilla convolutional autoencoder network whose encoder has a down-sampling rate of 16. \textbf{AE+Skip}: \textbf{AE} with skip connections. \textbf{MemAE}: \textbf{AE} augmented with the memory module proposed by~\cite{gong2019memorizing} and \{$2\times$, $4\times$, $8\times$, $16\times$\} indicate the down-sampling rates in our reimplementation. \textbf{MemAE+Skip}: \textbf{AE+Skip} equipped with the memory module~\cite{gong2019memorizing} in each skip connection.}
	\label{fig0}
\end{figure}

Reconstruction-based methods have been widely used for unsupervised anomaly detection. Most recent methods are based on convolutional neural networks (CNNs). In particular, autoencoder (AE)~\cite{kingma2013auto} is a natural choice to model the high-dimensional image data in the unsupervised setting. It is generally assumed~\cite{zong2018deep, hasan2016learning, zhao2017spatio} that the reconstruction error of the normal samples which are similar to the training data will be lower than abnormal samples.
However, due to the existence of down-sampling in AE, the latent coding representation loses the detailed information of the original image. It may result in blurry output and large reconstruction error even for normal samples.
We may strengthen the reconstruction capability of AE by reducing the down-sampling rate or adding skip connections~\cite{akccay2019skip}. But due to the uncontrollable generalizability of deep neural networks, it is difficult to improve the reconstruction quality of normal samples on one hand, and suppress the quality of abnormal samples on the other hand (see Figure~\ref{fig0}). Thus, the vanilla AE does not work well.

To enlarge the gap of reconstruction error between the normal and abnormal samples, a few works~\cite{gong2019memorizing,park2020learning} have explored the memory module. In particular, an external memory module is utilized to explicitly model the distribution of normality. However, according to our investigation, the gap of reconstruction error can not be effectively enlarged in the existing works, leading to limited performance improvement. As shown in Figure~\ref{fig0}, on the challenging MVTec AD~\cite{bergmann2019mvtec} dataset, when the down-sampling rate is large (e.g. $16\times$), MemAE~\cite{gong2019memorizing} failed to reconstruct both abnormal and some normal regions in the image. When the down-sampling rate is small (e.g. $2\times$) or with skip connections added, it reconstructed both normal and abnormal regions perfectly.

\begin{figure*}
	\centering
	\includegraphics[width=\linewidth]{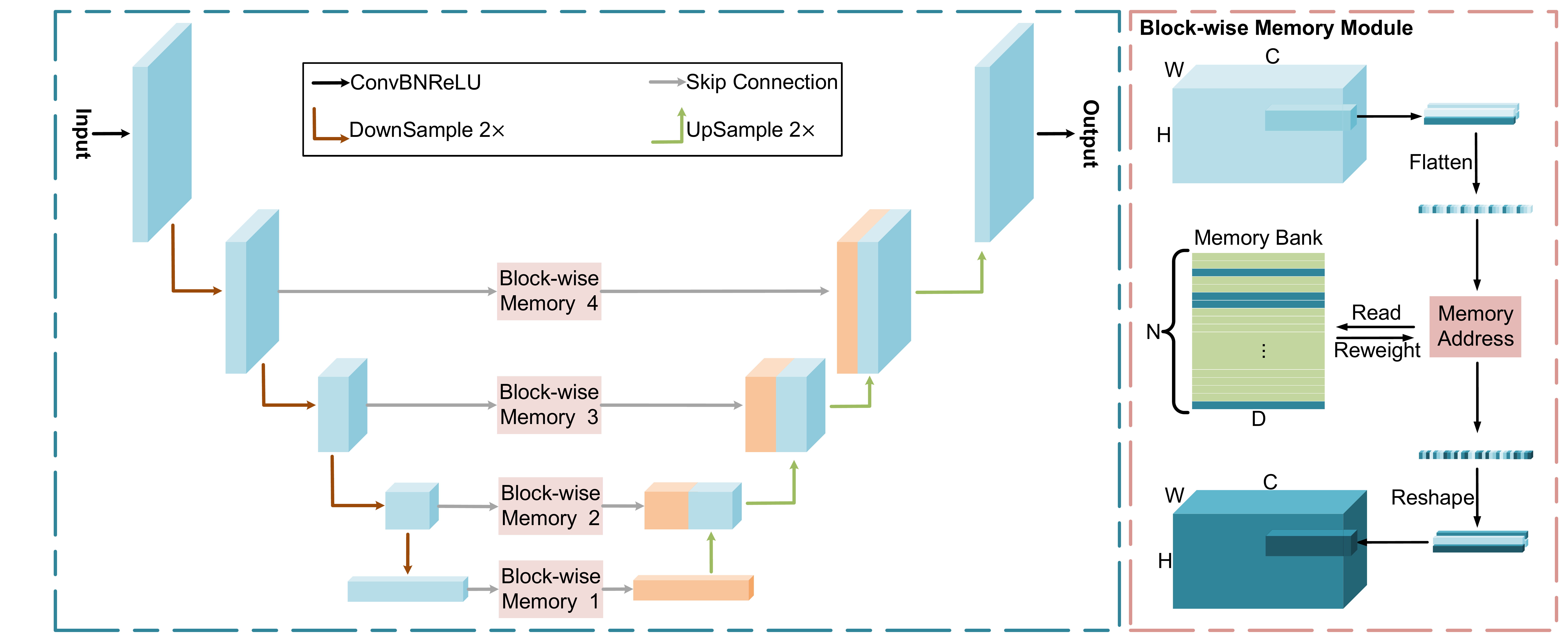}
	\caption{The architecture of the proposed DAAD framework. It consists of three major components: an encoder, a decoder and a group of block-wise memory modules. The encoder and decoder are connected with skip connections at multiple scales. And each skip connection is equipped with the block-wise memory module shown on the right.}
	\label{fig1-1}
\end{figure*}

We argue that the major weakness of existing memory-augmented models lies in the fact that they decode the encoded feature maps in a per-pixel manner. If the basic building block of feature maps is as small as one pixel, then the variation of patterns of each block is limited. The smaller the block is, the more likely that anomaly shares the same block patterns with normality and thus can be reconstructed accurately. This hypothesis can be conceptually explained with a toy example shown in Figure~\ref{toy-exp}. In this example, we try to learn block patterns of different sizes from normality A, and reconstruct normality B and anomaly C using the learned blocks. When the block is small, we learn two simple patterns (i.e. white and dark blue), and succeed to reconstruct both normality B and anomaly C. When the block size is too large, we fail to reconstruct both B and C. With a proper medium block size, we are able to reconstruct perfectly for normality B and poorly for anomaly C, which is ideal for a reconstruction-based anomaly detection system.

Following the above motivation, we propose the Divide-and-Assemble Anomaly Detection (DAAD) framework. Specifically, we interpret the reconstruction of an image as a divide-and-assemble procedure. The encoded feature maps are divided evenly into a grid of blocks. Accordingly, a block-wise memory module which matches with the size of the division is introduced to model the block-level rather than pixel-level distribution of normal samples. This design makes it possible to modulate the reconstruction capability of the model. And thus the tradeoff between good reconstruction on normal samples and poor reconstruction on abnormal samples can be achieved. Furthermore, skip connections can be added safely to improve the reconstruction quality of normal samples without reconstruction of abnormal samples. Given an input image, DAAD uses the multi-scale features of multiple encoding layers as queries to retrieve the most relevant items in the corresponding memory banks. Then those items are aggregated and sent to the decoder through the corresponding skip connections. The memory items are learned end-to-end together with the model parameters via backpropagation. As shown in Figure~\ref{fig0}, DAAD exhibits the desirable property of reconstructing details of normal regions while suppressing the reconstruction of abnormal regions.

In the real world, most anomalies such as scratch and crack tend to be subtle and occupy an extremely small region. Therefore even in abnormal images, the pixels are overwhelmed by normality rather than anomaly. For reconstruction-based methods, the cumulative error from normal pixels contributes more than abnormal pixels, which weakens the discriminativeness of the aggregated anomaly score. Even though the per-pixel reconstruction error of normal pixels is smaller than abnormal pixels, it is still difficult to detect anomaly. Thus we introduce an adversarially learned discriminator and exploit its low-dimensional semantic representation. The similarity between the features of the input image and the reconstructed image is treated as a measure of normality. The semantic feature-level anomaly score is complementary with the low-level reconstruction-based score, and their combination further improves the detection performance.

The major contributions of this paper are summarized as follows:

\begin{itemize}
	\item We propose the multi-scale block-wise memory module which is able to achieve a tradeoff between good reconstruction on normality and poor reconstruction on anomaly and thus make them well separable.
	\item We introduce an adversarially learned feature representation that detects anomaly in the semantic space and complements reconstruction-based detection.
	\item Extensive experiments on common unsupervised anomaly detection benchmarks demonstrate the excellent reconstruction and state-of-the-art performance.
\end{itemize}
\section{Related Work}
\textbf{Unsupervised Learning for Anomaly Detection.} Due to the scarcity and unpredictability of anomaly, anomaly detection is widely formulated as an unsupervised learning problem, where abnormal samples are unavailable during training~\cite{chandola2009anomaly}. Hence, the one-class classification methods such as one-class SVM~\cite{chen2001one}, support vector data descriptor~\cite{tax2004support} and deep one-class networks~\cite{ruff2018deep, chalapathy2018anomaly} are naturally chosen to solve this problem. Besides, unsupervised clustering methods (e.g. k-means method~\cite{zimek2012survey} and Gaussian Mixture Models~\cite{xiong2011group}) have also been explored to model the behavior of normal samples in the latent space for separating anomaly from normality. However, these methods are usually found difficult to obtain effective feature representation, resulting in sub-optimal performance when faced with high-dimensional image data.

\emph{Reconstruction-based methods} are mainly based on the assumption that a model trained only on normal samples can not represent and reconstruct anomaly accurately so that anomaly could be detected via the relatively higher reconstruction error than normality~\cite{zong2018deep}. Sakurada et al.~\cite{sakurada2014anomaly} applied autoencoder to anomaly detection for the first time. Later, Nicolau et al.~\cite{nicolau2016hybrid} proposed to take low density in the hidden layer to indicate anomaly by combining autoencoder and density estimation. To push the model to model the normality more effectively, recent works introduced more loss functions and network architectures. Sabokrou et al.~\cite{sabokrou2018adversarially} optimized the autoencoder with adversarial learning that could enhance the normal samples and distort the abnormal samples. Zenati et al.~\cite{zenati2018efficient} trained a BiGAN model and employed an external encoder to learn the reverse process of the generator. Furthermore, Akcay et al.~\cite{akccay2019skip} encoded the input and the reconstruction to the low-dimensional latent space using the discriminator to calculate the reconstruction error. Similarly, Wang et al.~\cite{wang2020advae} assumed that normal and abnormal data follow different Gaussian distributions and trained a variational autoencoder adversarially. Perera et al.~\cite{perera2019ocgan} proposed the OCGAN that embeds two discriminators and a classifier on a denoising autoencoder.
To suppress the reconstruction of anomaly, ITAE~\cite{huang2019inverse} and AESc~\cite{collin2020improved} take the idea of image restoration. ITAE erases selected attributes (color or rotation) while AESc corrupts the original clean normality to simulate anomaly. Then they try to restore the corruptions in the training stage. Our work belongs to the reconstruction-based method. By extending the vanilla AE with the multi-scale block-wise memory module and the adversarially learned representation, we improve the baseline model significantly.

\emph{Feature-based methods} exploit pre-trained deep representations in a transfer learning setting to model the distribution of normality. Recently, a few works such as SPADE~\cite{cohen2020sub}, MAHA~\cite{rippel2020modeling} and FAVAE~\cite{dehaene2020anomaly} attempt to exploit the powerful knowledge learned from an external large-scale dataset. Specifically, a classification model is pre-trained on the ImageNet dataset, which is subsequently used as a fixed feature extractor. Then SPADE and MAHA identify anomaly by comparing the semantic features of a test sample with a set of normal samples. FAVAE uses the pre-trained model to guide the training of a variational auto-encoder. Our approach does not make use of any external knowledge, which makes the comparison unfair. Nevertheless, we compare their performance as a reference, and our approach still surpasses most feature-based methods.

\textbf{Memory Networks.} In order to capture long-term dependency in sequential data, recurrent neural network (RNN) and long short-term memory (LSTM)~\cite{hochreiter1997long} were proposed. They use hidden states of the network to selectively record past information as their memory. But sometimes they fail to work well since the memory capacity is too small to accurately record all the contents of the sequential data. Recently, Weston et al.~\cite{weston2014memory} introduced the memory networks that use a specialized memory bank that can be read and written and perform better memorization. However, it is hard to train the memory network via backpropagation due to the need of supervision for each layer during training. Sukhbaatar et al.~\cite{sukhbaatar2015end} made it possible to train the memory networks in an end-to-end manner by using continuous memory representation to read/write the memory. Based on the end-to-end memory network, Miller et al.~\cite{miller2016key} utilized key-value pairs to further extend the memory capability. The success and effectiveness of the memory networks have attracted increasing interests for different computer vision tasks including object tracking~\cite{yang2018learning, lai2020mast, marchetti2020mantra}, action recognition~\cite{yuan2019memory}, visual question answering~\cite{yin2019memory, fan2019heterogeneous} and anomaly detection~\cite{gong2019memorizing, park2020learning, zhang2020memory, yang2020memory}. Similar to~\cite{gong2019memorizing, park2020learning}, we also explore the potential of memory module in the context of anomaly detection. Our major new contribution is to modulate the reconstruction capability on normality and anomaly by generalizing the memory module to be block-wise rather than pixel-wise in a multi-scale fashion.

\section{Approach}
In this section, we first outline the proposed DAAD framework for unsupervised anomaly detection. Then, we present the detailed components of the block-wise memory module. Next, we introduce an adversarially learned feature representation that assists to identify anomaly by comparing the features of the input image and the reconstructed image. Finally, we describe the training loss function and the way to compute the anomaly score.

\subsection{Network Architecture}

\subsubsection{Encoder and Decoder}

The left part of Figure~\ref{fig1-1} shows the architecture of the proposed memory-augmented autoencoder network. It adopts a bow-tie network with an encoder $G_E$ and a decoder $G_D$. The encoder network maps an input image $x$ into a latent representation $z$. Being symmetric to $G_E$, the decoder network $G_D$ upsamples the latent vector $z$ back to the input resolution and predicts the reconstruction, denoted as $\hat{x}$. Inspired by the UNet architecture~\cite{ronneberger2015u} which is widely used for dense prediction tasks, we insert skip connections at multiple scales between the encoder and decoder. Using skip connections provides substantial advantages via direct information transfer between the layers. It can preserve both local and global information, and hence yield better reconstruction. However, it also brings a high risk of learning an identity mapping from input to output, which would make anomaly and normality inseparable as demonstrated in Figure~\ref{fig0}. To address this issue, we equip all connections between the encoder and decoder with the block-wise memory module, which is described in the following section.

\subsubsection{Block-wise Memory Module}

The divide-and-assemble concept is effectively implemented using the block-wise memory module. It involves the block-wise query representation and a memory bank to record prototypical patterns. An attention-based addressing operator is designed for access and updating of the memory items.

\paragraph{Block-wise Query Representation.}
Given the input image $x$, the encoder emits a set of feature maps $\{f_1, f_2, \dots, f_L\}$ where $L$ is the number of scales. For each feature map $f_l$ of size $H^l \times W^l \times C^l$, we divide it evenly into a grid of $Q=r_h\times r_w \times r_c$ blocks, where $r_h$, $r_w$ and $r_c$ are the division rates along height, width and channels respectively. The blocks in the grid are denoted as $\{\mathbf{q}^k \in \mathbb{R}^{\frac{H^l}{r_{h}}\times \frac{W^l}{r_{w}} \times \frac{C^l}{r_{c}}}, k=1, 2, \ldots, Q\}$. Each block is flattened into a vector and serves as a query to the memory bank. For each query, unique features are aggregated from the memory bank with the memory addressing operation. The aggregated features share the same dimension as the query and are reshaped back into a block of $\frac{H^l}{r_{h}}\times \frac{W^l}{r_{w}} \times \frac{C^l}{r_{c}}$. By assembling the aggregated feature blocks, we obtain the transformed feature maps $\{\hat{f}_1, \hat{f}_2, \dots, \hat{f}_L\}$, which are subsequently fed into the decoder $G_D$ for the reconstruction of $x$. Note that we use the same set of division rates ($r_h$, $r_w$ and $r_c$) for all scales, which results in different block sizes for feature maps of different levels. And the queries at different levels do not share the same memory bank.

\paragraph{Memory Bank and Addressing.}
The memory bank $\mathbf{M} \in \mathbb{R}^{N \times D}$ is defined as a real-valued matrix containing $N$ memory items of fixed dimension $D$. It is used to explicitly record the prototypical normal patterns during training. To facilitate calculation, we set $D$ equal to the dimension of the query $\mathbf{q}^{k}$. The $i$-th row vector of $\mathbf{M}$ denotes one memory item $\mathbf{m}_{i} \in \mathbb{R}^{D}$ ($i \in \{1, 2, \dots, N\}$). Similar to~\cite{weston2014memory, rae2016scaling}, the memory is content addressable which has a specific addressing scheme. It is addressed by computing the attention weights $\mathbf{w}$ based on the similarity between the query $\mathbf{q}^k$ and each item $\mathbf{m}_i$ in the memory bank. Concretely, each weight $w_{i}$ is computed as follows:
\begin{equation}
\label{eq1}
w_{i}=\frac{\exp \left(\frac{\mathbf{q}^{k} \mathbf{m}_{i}^{\mathsf T}}{\|\mathbf{q}^{k}\|\left\|\mathbf{m}_{i}\right\|}\right)}{\sum_{j=1}^{N} \exp \left(\frac{\mathbf{q}^{k} \mathbf{m}_{j}^{\mathsf T}}{\|\mathbf{q}^{k}\|\left\|\mathbf{m}_{j}\right\|}\right)}
\end{equation}
Note that $w_i$ is non-negative and all elements in $\mathbf{w} \in \mathbb{R}^{1 \times N}$ summate to one. Based on the attention weights $\mathbf{w}$, a new feature block $\hat{\mathbf{q}}^{k}$ is aggregated over the memory bank by
\begin{equation}
\hat{\mathbf{q}}^{k}=\mathbf{w} \mathbf{M}=\sum_{i=1}^{N} w_{i} \mathbf{m}_{i}
\label{eq3}
\end{equation}

As shown in Eq.~\eqref{eq1} and \eqref{eq3}, the addressing weights $\mathbf{w}$ is used to retrieve the memory items most related to $\mathbf{q}^{k}$ and generate the corresponding normal representation $\hat{\mathbf{q}}^{k}$. By linear combination over the memory items, some abnormal images may still achieve good reconstruction. To alleviate this problem, we drop the items whose memory weights are less than $\frac{1}{N}$ when computing $\hat{\mathbf{q}}^{k}$. Due to the sparsity in $\mathbf{w}$, only a small number of memory items can be addressed each time.

\subsubsection{Adversarially Learned Representation}
\begin{figure}
	\centering
	\includegraphics[width=\linewidth]{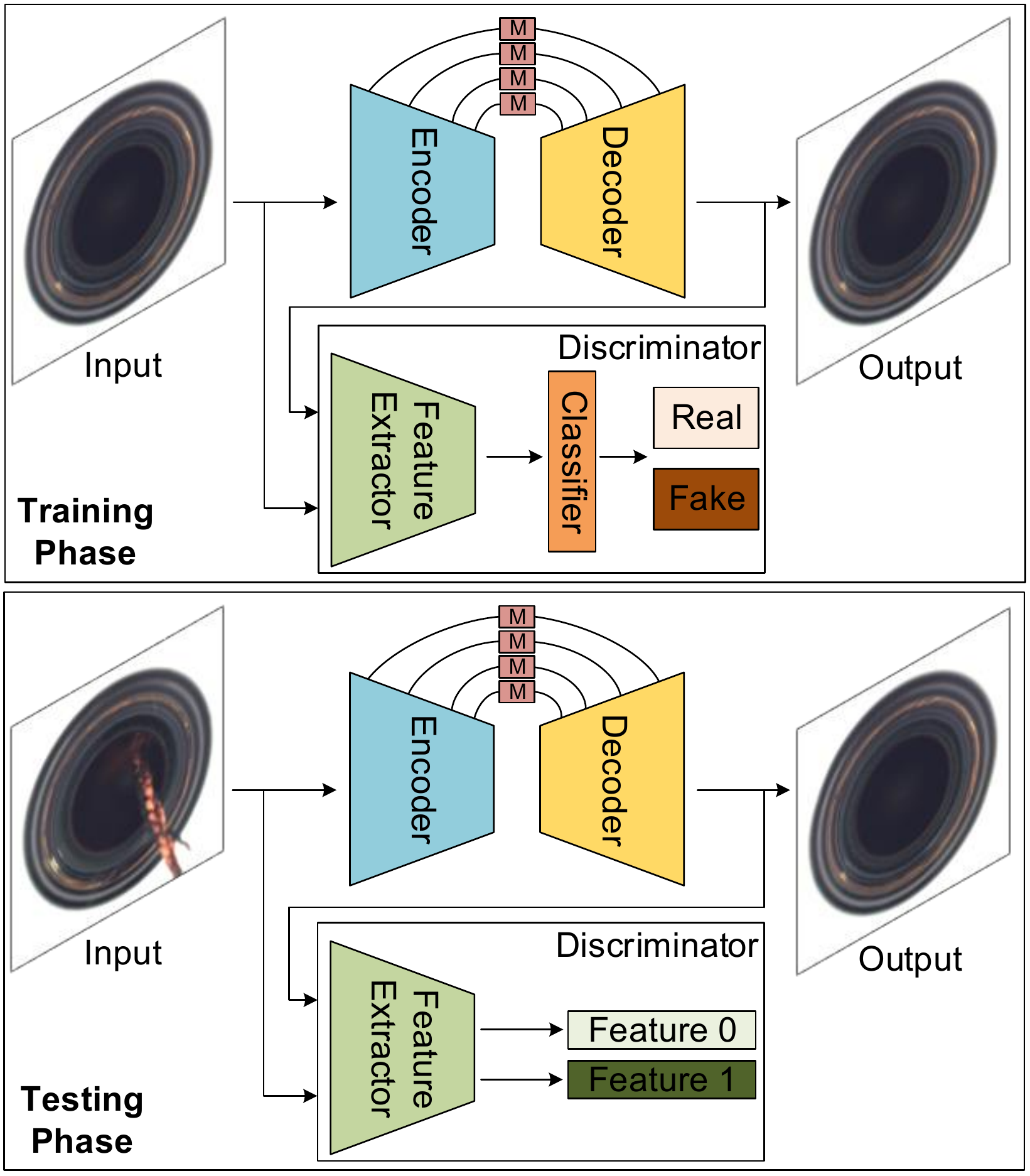}
	\caption{Diagram of the adversarially learned representation. A discriminator is appended after the decoder and the encoder-decoder network acts as the generator.}
	\label{fig2}
\end{figure}

For some complex images that are difficult to reconstruct, the anomaly score based on the reconstruction loss easily suffers from the cumulative error of a large number of normal pixels. To address this issue, we introduce an adversarially learned representation. During training, we minimize the distance between the original image and the reconstructed image in the low-dimensional semantic representation space. During testing, we incorporate the feature similarity into the anomaly score to make it more robust in the case of poor reconstruction of normality. To prevent the learned features from mode collapse~\cite{grill2020bootstrap, chen2020exploring}, we utilize the adversarial training strategy. Specifically, the encoder-decoder network is treated as a generator $G$, and a discriminator $D$ is appended after the decoder. As shown in Figure~\ref{fig2}, the discriminator predicts the real/fake label of the given input. Its architecture is designed by following the discriminator of DCGAN~\cite{radford2015unsupervised}. During training, $D$ tries to differentiate real images $x$ from the fake ones $\hat{x}$ generated by $G$. During inference, the network $D$ is used as a feature extractor to extract features of the input image $x$ and the reconstructed image $\hat{x}$.

\subsection{Loss Function}

The overall training loss is a weighted summation of three losses:
\begin{equation}
\label{eq4}
\mathcal{L}=\lambda_{rec} \mathcal{L}_{rec}+\lambda_{adv} \mathcal{L}_{adv}+\lambda_{ali} \mathcal{L}_{ali}
\end{equation}
where $\mathcal{L}_{rec}$, $\mathcal{L}_{adv}$ and $\mathcal{L}_{ali}$ are the reconstruction, adversarial and alignment losses respectively. $\lambda_{rec}$, $\lambda_{adv}$ and $\lambda_{ali}$ are the corresponding loss weights.

\begin{table*}[]
	\footnotesize
	\setlength{\tabcolsep}{3pt}
	\begin{center}
		\resizebox{\textwidth}{!}{%
			\begin{tabular}{c|l|ccccccccccccccc|c}
				\hline
				\multicolumn{2}{c|}{Category}                        & Carpet & Grid  & Leather & Tile  & Wood  & Bottle & Cable & Capsule & HN    & MN    & Pill  & Screw & TB    & TS    & Zipper & Mean           \\ \hline
				\multirow{3}{*}{Feature}        & SPADE~\cite{cohen2020sub}       & -      & -     & -       & -     & -     & -      & -     & -       & -     & -     & -     & -     & -     & -     & -      & 0.855          \\
				& MAHA~\cite{rippel2020modeling} & 1.000      & 0.897 & 1.000       & 0.998 & 0.996 & 1.000      & 0.950  & 0.951   & 0.991 & 0.947 & 0.887 & 0.852 & 0.969 & 0.955 & 0.979  & \textbf{0.958}  \\
				& FAVAE~\cite{dehaene2020anomaly}       & 0.671  & 0.970  & 0.675   & 0.805 & 0.948 & 0.999  & 0.950  & 0.804   & 0.993 & 0.852 & 0.821 & 0.837 & 0.958 & 0.932 & 0.972  & 0.879          \\ \hline
				\multirow{5}{*}{Recons.} & GeoTrans$^\ast$~\cite{golan2018deep}    & 0.437  & 0.619 & 0.841   & 0.417 & 0.611 & 0.744  & 0.783 & 0.670    & 0.359 & 0.813 & 0.630  & 0.500   & 0.972 & 0.869 & 0.820   & 0.672          \\
				& GANomaly$^\ast$~\cite{akcay2018ganomaly}    & 0.699  & 0.708 & 0.842   & 0.794 & 0.834 & 0.892  & 0.757 & 0.732   & 0.785 & 0.700   & 0.743 & 0.746 & 0.653 & 0.792 & 0.745  & 0.762          \\
				& ITAE~\cite{huang2019inverse}        & 0.706  & 0.883 & 0.862   & 0.735 & 0.923 & 0.941  & 0.832 & 0.681   & 0.855 & 0.667 & 0.786 & 1.000     & 1.000     & 0.843 & 0.876  & 0.839          \\
				& AESc~\cite{collin2020improved}        & 0.89   & 0.97  & 0.89    & 0.99  & 0.95  & 0.98   & 0.89  & 0.74    & 0.94  & 0.73  & 0.84  & 0.74  & 1.00     & 0.91  & 0.94   & 0.89          \\
				& Ours        & 0.866  & 0.957 & 0.862   & 0.882 & 0.982 & 0.976  & 0.844 & 0.767   & 0.921 & 0.758 & 0.900   & 0.987 & 0.992 & 0.876 & 0.859  & \textbf{0.895} \\ \hline
			\end{tabular}%
		}
	\end{center}
	\caption{Comparison with state-of-the-art feature-based and reconstruction-based methods on the MVTec AD dataset in terms of the image-level AUROC score. For clarity, the category names Hazelnut, Metal Nut, Toothbrush and Transistor are abbreviated as HN, MN, TB and TS respectively. $^\ast$ indicates reproduced results by~\cite{huang2019inverse}.}
	\label{table1}
\end{table*}

\paragraph{Reconstruction Loss.} The reconstruction loss makes the output image reconstructed from the generator $G$ similar to its input image. Specifically, we apply $L_2$ distance between the input $x$ and the reconstructed output $\hat{x}$, which ensures that our model is capable of reconstructing images contextually similar to normal images. The reconstruction loss is shown below:
\begin{equation}
\label{eq5}
\mathcal{L}_{\text {rec}}=\underset{x \sim p_{x}}{\mathbb{E}}||x-\hat{x}||_{2}
\end{equation}
Here, $p_{x}$ denotes the distribution of normal samples.

\paragraph{Adversarial Loss.} To train the discriminator $D$ for adversarially learned representation, we utilize the adversarial loss proposed in~\cite{goodfellow2014generative}. As shown in Eq.~\eqref{eq6}, the discriminator $D$ tries to classify the real (input) and fake (reconstructed) samples correctly and the generator $G$ tries to reconstruct $x$ as realistically as possible. The adversarial loss $\mathcal{L}_{adv}$ can be written as:
\begin{equation}
\label{eq6}
\mathcal{L}_{adv}=\underset{x \sim p_{x}}{\mathbb{E}}[\log D(x)]+\underset{x \sim p_{x}}{\mathbb{E}}[\log (1-D(G(x))]
\end{equation}

\paragraph{Alignment Loss.} The alignment loss aims to match the feature representations of the reconstructed image $\hat{x}$ to the input image $x$. It helps to ensure that the feature similarity is informative to indicate anomaly, i.e. higher for normal samples than abnormal samples. As shown in Figure~\ref{fig2}, by feeding a pair of images ($x$ and $\hat{x}$) into the discriminator D, we extract their features at the last convolutional layer of D. The extracted features $f_D(x)$ and $f_D(\hat{x})$ serve as semantic representations of $x$ and $\hat{x}$. The alignment loss therefore is defined as follows:
\begin{equation}
\label{eq7}
\mathcal{L}_{ali}=\underset{x \sim p_{x}}{\mathbb{E}}||f_D(x)-f_D(\hat{x})||_{2}
\end{equation}

\subsection{Anomaly Score}
During inference, anomaly is detected by fusing the reconstruction error in the high-dimensional image space and the alignment error in the low-dimensional feature space between the input image and the reconstructed image. Given the input image $x$, we define the anomaly score as follows:
\begin{equation}
\label{eq8}
\mathcal{A}(x)=\gamma R(x)+(1-\gamma) L(x)
\end{equation}
where $R(x)$ is the reconstruction score measuring the average pixel-wise difference between the input and the reconstructed images based on Eq.~\eqref{eq5}. $L(x)$ is the alignment score measured by the distance between the feature representations of the input and the reconstructed images based on Eq.~\eqref{eq7}. $\gamma$ is a weighting parameter controlling the contribution of the reconstruction score and the alignment score. Note that both $R(x)$ and $L(x)$ are linearly scaled to the range of $[0, 1]$ before fusion.

\section{Experiments}

\subsection{Setup}
This section introduces the datasets, implementation details and the evaluation criterion used in our experiments.

\paragraph{Datasets.}
To demonstrate the proof of concept of the proposed approach, we validate the model on the challenging industrial anomaly detection dataset MVTec AD~\cite{bergmann2019mvtec}. The MVTec AD dataset contains 5354 high-resolution color images of 15 different classes. 5 classes consist of textures such as wood or leather. The other 10 classes contain objects. Different from those used in the existing benchmarks, e.g. MNIST~\cite{lecun1998mnist} and CIFAR10~\cite{krizhevsky2009learning}, the anomalies in this dataset are more fine-grained and abnormal images come from the same category, which makes it more challenging. The dataset contains a training set with only normal images and a testing set with both normal and abnormal images. 

\paragraph{Implementation Details.}
We resize each input image into a size of $256 \times 256$ for MVTec AD, and normalize the pixel values to the range of $[-1, 1]$. We use horizontal and vertical image flipping for data augmentation. Unless otherwise noted, $r_{h}, r_{w}, r_{c}$ and the memory bank size $N$ are empirically set to 8, 8, 1 and 500 to achieve the best overall performance. When only using the reconstruction loss (DAAD), the model is optimized via the Adam~\cite{kingma2014adam} optimizer with a fixed learning rate of $1 e^{-4}$, a weight decay of 0, and momentums of $\beta_{1}=0.9$, $\beta_{2}=0.999$. If the adversarial loss and alignment loss are enabled (DAAD+), the learning rate is increased to $2 e^{-4}$. The weighting parameters of $\mathcal{L}$ are empirically chosen as $\lambda_{rec}=50$, $\lambda_{adv}=0.5$ and $\lambda_{ali}=0.5$. When applicable, the $\gamma$ in Eq.~\eqref{eq8} is set to 0.9. The model is implemented using PyTorch~\cite{paszke2017automatic} and trained for 1000 epochs with a batch size of 8.

\paragraph{Evaluation.}
The performance of the model is evaluated by the area under the curve (AUC) of the receiver operating characteristics (ROC) at the image level (AUROC), a function of true positive rate over false positive rate with varying thresholds.

\subsection{Results}

\paragraph{MVTec AD.}
On this dataset, we compare our approach with the existing state-of-the-art methods, including both feature-based and reconstruction-based methods. As shown in Table~\ref{table1}, our approach surpasses most reconstruction-based methods by a large margin. Our performance is comparable to AESc~\cite{collin2020improved} which exploits image restoration techniques by generating synthetic anomaly. We expect that these techniques are complementary to the proposed framework. Notably, although the comparison is unfair due to their usage of external knowledge, we still outperform most feature-based methods, i.e. SPADE~\cite{cohen2020sub} and FAVAE~\cite{dehaene2020anomaly}. Remarkably, compared with the vanilla AE, we obtain an absolute AUROC gain of 10.1\% as discussed in Section~\ref{sect:ablation-study}.

\paragraph{MNIST and CIFAR10.}
To validate the versatility of the proposed framework, we report the performance on the commonly used MNIST and CIFAR10 datasets. We follow the one-class novelty detection protocol used in~\cite{perera2019ocgan}, and the comparison with existing reconstruction-based methods is listed in Table~\ref{table4}. Our approach also achieves excellent performance.

\begin{table}[t]
	\begin{center}
		\begin{threeparttable}
			\begin{tabular}{l|cc}
				\hline
				Method  & MNIST & CIFAR10 \\ \hline
				VAE~\cite{kingma2013auto}      & 0.877  & 0.581    \\
				AnoGAN~\cite{schlegl2017unsupervised}   & 0.937  & 0.612    \\
				ADGAN~\cite{deecke2018image}    & 0.947  & 0.624    \\
				GANomaly~\cite{akcay2018ganomaly} & 0.928  & 0.695    \\
				GeoTrans~\cite{golan2018deep} & 0.980  & 0.656    \\
				Skip-GANomaly~\cite{akccay2019skip}     & -  & 0.732    \\				
				OCGAN~\cite{perera2019ocgan}    & 0.975  & 0.656    \\
				MemAE\tnote{1} ~\cite{gong2019memorizing}   & 0.975 & 0.609   \\ 
				CAVGA-D$_{u}$~\cite{2020Attention}    & 0.986  & 0.737    \\ \hline
                Ours     & \textbf{0.990}  & \textbf{0.753}   \\ \hline
			\end{tabular}
			\begin{tablenotes}
				\footnotesize
				\item[1] MemAE uses a different evaluation protocol, where only part of the test set are used.
			\end{tablenotes}
		\end{threeparttable}
	\end{center}
	\caption{Comparison with existing methods on the MNIST and CIFAR10 datasets in terms of the AUROC score.}
	\label{table4}
\end{table}

\subsection{Ablation Study}
\label{sect:ablation-study}

To validate the effectiveness of the individual components of the proposed framework, we perform an extensive ablation study on the MVTec AD dataset.

\paragraph{Improvement over Baseline.}
We improve the vanilla AE and existing techniques like the memory module in~\cite{gong2019memorizing} and skip connections in~\cite{akccay2019skip} significantly. The detailed comparison is shown in Table~\ref{table2}. AE is a vanilla autoencoder network trained with the reconstruction loss. MemAE is AE augmented with the pixel-wise memory module proposed in~\cite{gong2019memorizing}, which is re-implemented by using a single-scale ($16\times$) block-wise memory module with $r_{h} = 16$, $r_{w}=16$ and $r_{c}=1$. AE+Skip is AE equipped with skip connections. DAAD is AE equipped with the proposed multi-scale block-wise memory module. DAAD+ is DAAD complemented with the adversarially learned representation. Taking AE as a baseline model, MemAE barely improves AE, indicating that the pixel-wise memory module does not work well for complex scenes like MVTec AD. AE+Skip fails to improve AE as it improves the reconstruction quality of both normality and abnormality, which makes them inseparable. We notice that AE+Skip can get good performance on some categories, e.g. tile, wood and screw. That is because the foreground and background texture of these categories are relatively homogeneous, which makes the reconstruction of anomaly more distinct. Nevertheless, compared with AE, DAAD achieves an AUROC gain of $5.1\%$, verifying the effectiveness of the proposed multi-scale block-wise memory module. DAAD+ further improves AUROC by 5\%, resulting in a total improvement of 10.1\%, which proves that the adversarially learned representation is complementary with DAAD.

\begin{table}[t]
	\begin{center}
		\small
		\setlength{\tabcolsep}{4pt}
		\begin{tabular}{c|ccccc}
			\hline
			Category   & AE    & MemAE$^\star$ & AE+Skip  & DAAD & DAAD+ \\ \hline
			Carpet     & 0.411 & 0.454 & 0.385 & 0.671    & 0.866 \\
			Grid       & 0.841 & 0.946 & 0.879 & 0.975   & 0.957 \\
			Leather    & 0.615 & 0.611 & 0.570  & 0.628   & 0.862 \\
			Tile       & 0.696 & 0.630  & 0.986 & 0.825   & 0.882 \\
			Wood       & 0.961 & 0.967 & 0.977 & 0.957   & 0.982 \\
			Bottle     & 0.955 & 0.954 & 0.713 & 0.975   & 0.976 \\
			Cable      & 0.688 & 0.694 & 0.579 & 0.720    & 0.844 \\
			Capsule    & 0.819 & 0.831 & 0.747 & 0.866   & 0.767 \\
			Hazelnut   & 0.884 & 0.891 & 0.828 & 0.893   & 0.921 \\
			Metal Nut  & 0.565 & 0.537 & 0.336 & 0.552   & 0.758 \\
			Pill       & 0.882 & 0.883 & 0.853 & 0.898   & 0.900 \\
			Screw      & 0.956 & 0.992 & 1.000 & 1.000   & 0.987 \\
			Toothbrush & 0.977 & 0.972 & 0.742 & 0.989   & 0.992 \\
			Transistor & 0.776 & 0.793 & 0.749 & 0.814   & 0.876 \\
			Zipper     & 0.878 & 0.871 & 0.696 & 0.906   & 0.859 \\ \hline
			Mean       & 0.794 & 0.802 & 0.736 & 0.845   & \textbf{0.895} \\ \hline
		\end{tabular}
	\end{center}
	\caption{AUROC improvement on MVTec AD over the vanilla AE and existing techniques. $^\star$ denotes our reproduced results.}
	\label{table2}
\end{table}

\begin{table}[t]
	\begin{center}
		\begin{tabular}{c|ccccc}
			\hline
			$r_h$ \& $r_w$ & 1 & 2 & 4 & 8 & 16 \\\hline
			AUROC & 0.716 & 0.784 & 0.821 & \textbf{0.845} & 0.813 \\
			\hline
		\end{tabular}
	\end{center}
	\caption{Impact of block size of the block-wise memory module on AUROC. Note that using a larger division rate results in a smaller block size.}
	\label{tab-block-size}
\end{table}

\begin{figure}
	\centering
	\includegraphics[width=\linewidth]{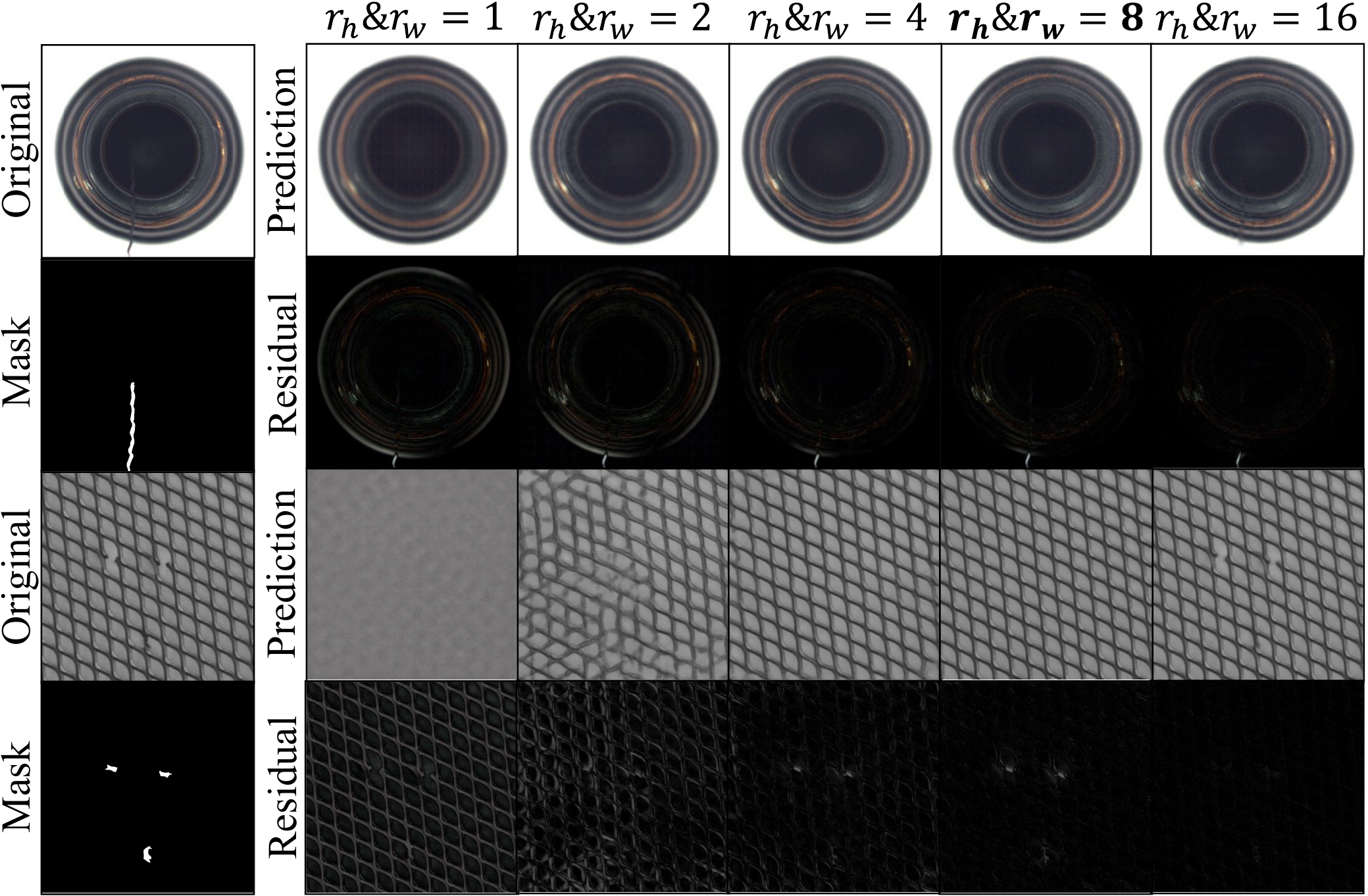}
	\caption{Two examples verifying that the reconstruction quality varies with the memory block size. The reconstructed images using a division rate ($r_{h}$ \& $r_{w}$) of 1, 2, 4, 8 and 16 are shown from left to right. As the division rate increases, the block size is smaller, and the reconstruction becomes more clear and even identical to the input image.}
	\label{fig3}
\end{figure}

\paragraph{Impact of Block Size.}
To validate our hypothesis on the impact of block size as demonstrated in Figure~\ref{toy-exp}, we evaluate the AUROC score using different division rates along the spatial dimensions ($r_h$ and $r_w$), which essentially determine the block size. As listed in Table~\ref{tab-block-size}, using either a small or large division rate leads to sub-optimal performance. While using a medium division rate (i.e. 8) achieves the best performance. In Figure~\ref{fig3}, we visualize the reconstruction of two typical images under the same set of division rates. We can find that the reconstruction quality is closely related to the block size regulated by the division rate. For the object category of bottle, when the block size is large (with a small division rate), the model can only reconstruct the approximate appearance of the input and lose a lot of detailed texture, including the abnormal pixels. When the block size becomes smaller, the model can gradually reconstruct the details of the image. And when the block size is small enough, the model can even reconstruct abnormal pixels. For the texture category of grid, when the block size is too large, the model fails to reconstruct the original image. When the block size becomes smaller, the reconstructed texture gradually becomes complete and even identical to the input image. Similar results can be observed by varying the division rate along channels ($r_c$). In summary, our motivation of the divide-and-assemble procedure is well supported by both quantitative and qualitative results. By choosing a proper division rate, we are able to achieve a tradeoff between good reconstruction for normal pixels and poor reconstruction for abnormal pixels.

\begin{table}[]
  \begin{center}
	\begin{tabular}{l|ccccc}
		\hline
		Method& $\mathcal{L}_{adv}$ & $\mathcal{L}_{ali}$ & $L(x)$ & $R(x)$ & AUROC                \\ \hline
		AE                    &                   &                 &              &$\checkmark$  & 0.794                \\
		& $\checkmark$                 &                 &              & $\checkmark$ & 0.812                    \\
		& $\checkmark$                 & $\checkmark$               &             & $\checkmark$&       0.817              \\
		& $\checkmark$                 & $\checkmark$               & $\checkmark$            &  &       0.819              \\
		\multicolumn{1}{l|}{} & $\checkmark$                 & $\checkmark$               &  $\checkmark$             &$\checkmark$ &0.835 \\ \hline
		MemAE$^\star$                 &                   &                 &           &$\checkmark$& 0.802                \\
		& $\checkmark$                 &                 &            &$\checkmark$& 0.825                     \\
		& $\checkmark$                 & $\checkmark$               &              &$\checkmark$& 0.835                     \\
		& $\checkmark$                 & $\checkmark$               & $\checkmark$            &  &       0.841              \\
		\multicolumn{1}{l|}{} & $\checkmark$                 & $\checkmark$               &  $\checkmark$       &$\checkmark$& 0.853\\ \hline
		DAAD                  &                   &                 &              &$\checkmark$& 0.845                \\
		& $\checkmark$                 &                 &              &$\checkmark$ &           0.847     \\
		\multicolumn{1}{l|}{} & $\checkmark$                 & $\checkmark$               &              &$\checkmark$& 0.855 \\
		& $\checkmark$                 & $\checkmark$               & $\checkmark$            &  &       0.874             \\
		DAAD+                 & $\checkmark$                 & $\checkmark$               & $\checkmark$              &$\checkmark$& 0.895                \\ \hline
	\end{tabular}%
  \end{center}
	\caption{Quantitative comparison for variants of adversarially learned representation on  MVTec AD dataset.}
\label{tab-adv-representation}
\end{table}

\begin{figure}[t]
	\centering
	\includegraphics[width=\linewidth]{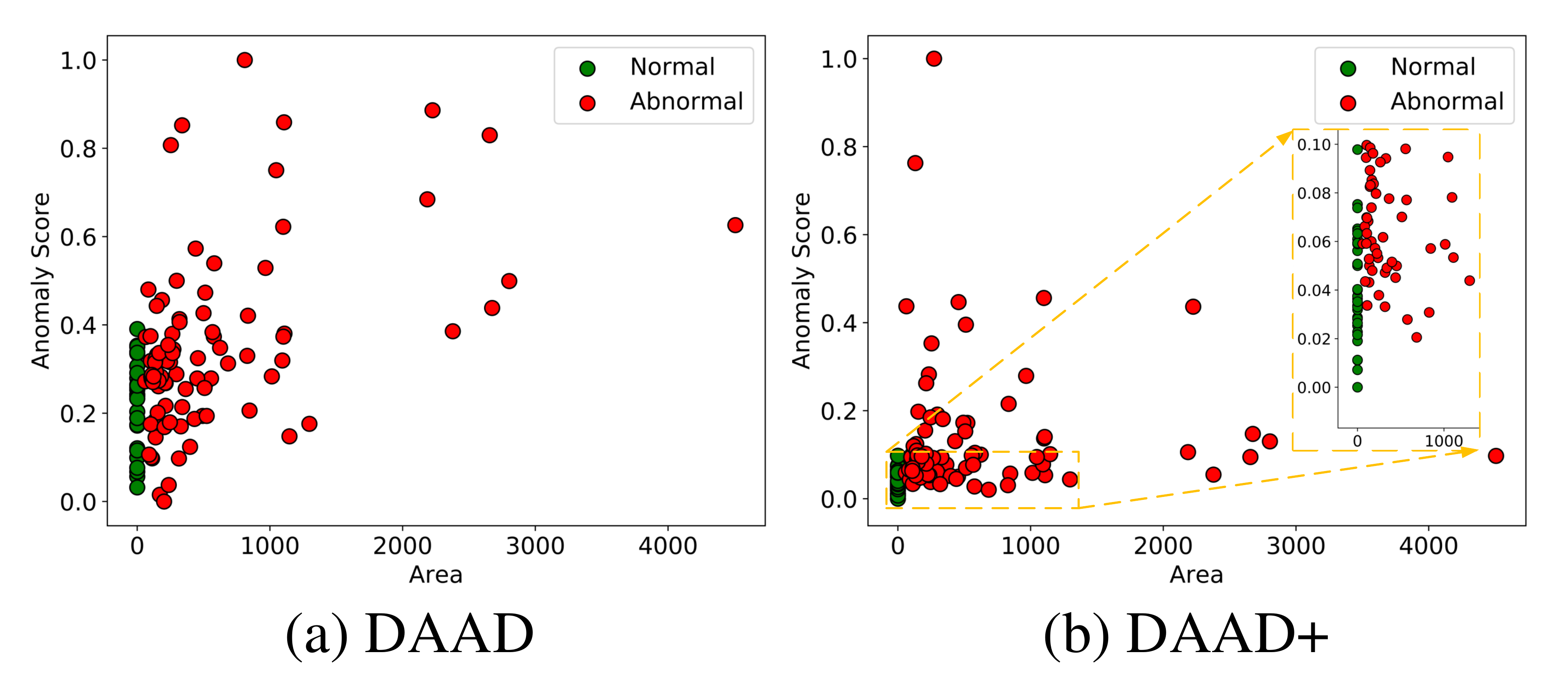}
	\caption{Scatter plot of the anomaly score and the anomaly area for the \emph{leather} class of the MVTec AD dataset.}
	\label{fig5}
\end{figure}

\paragraph{Impact of Adversarially Learned Representation.}
Table~\ref{tab-adv-representation} shows the effectiveness of the adversarially learned representation based on different generators, i.e. AE, MemAE and DAAD. We observe that adversarial training alone brings no performance gain for DAAD, which indicates that DAAD has been able to achieve a perfect reconstruction, and it is difficult to further improve the performance based on the reconstruction score only by the adversarial loss $\mathcal{L}_{adv}$. The alignment loss $\mathcal{L}_{ali}$ essentially adds an additional constraint to the model, thereby achieving a small gain. However, we obtain a performance boost by integrating the alignment score $L(x)$ into the anomaly score. 
We believe that the reconstruction score $R(x)$ based on the input image and the reconstructed image can only attend to large or obvious anomalies. While the alignment score calculated in a low-dimensional space can further suppress the anomaly score of normality, especially for categories that are difficult to reconstruct, such as leather and carpet. Figure~\ref{fig5} compares the distribution of anomaly score of DAAD and DAAD+, which clearly validates our argument.
\section{Conclusion}

As a challenging unsupervised learning problem, many techniques have been exploited in the context of anomaly detection.
The memory module perfectly fits the task of modeling the patterns of normality. Yet existing works failed to effectively improve the performance, especially on complex datasets like MVTec AD. In this work, we interpret the reconstruction of an image from the novel perspective of division and assembly of the feature maps. By controlling the granularity of the building blocks, i.e. the memory items, we achieve a balance between good reconstruction on normality and poor reconstruction on anomaly, which is the key to the success of reconstruction-based anomaly detection methods. Besides, using the adversarially learned feature representation, we are able to detect anomaly in the low-dimensional semantic space, which complements the detection in the image space. The excellent results on common anomaly detection benchmarks validate the effectiveness and versatility of the proposed framework.

{\small
	\bibliographystyle{ieee_fullname}
	\bibliography{egbib}
}

\clearpage
\section*{Appendix}

\begin{figure}[]
	\centering
	\includegraphics[width=\linewidth]{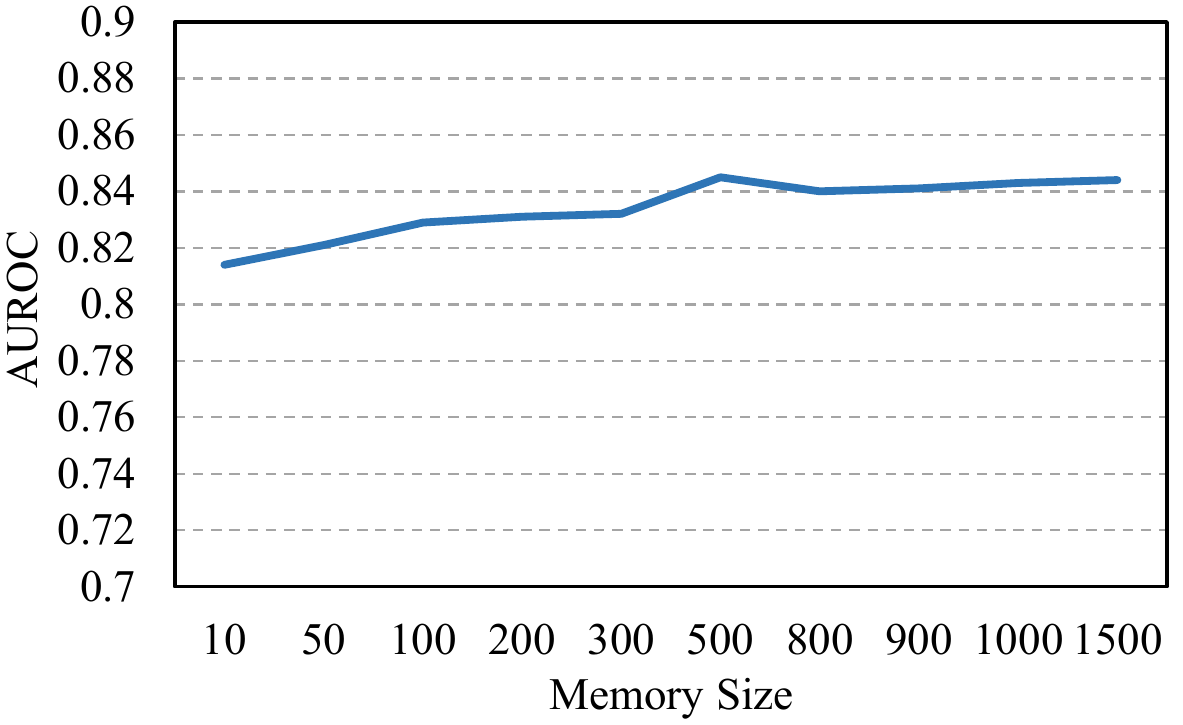}
	\caption{The AUROC score of DAAD over different memory sizes on the MVTec AD dataset.}
	\label{fig4}
\end{figure}

\paragraph{Impact of Memory Size.}
Figure~\ref{fig4} shows the impact of the memory size, i.e. the number of items $N$ in the memory bank to the detection performance. As the memory size increases, AUROC gets steadily improved, and saturates after 500. Notably, increasing the memory size further does not lead to an obvious performance degradation. It indicates that the proposed block-wise memory module enjoys the benefit of good reconstruction on normal samples without worrying about learning an identity mapping, which would not be possible by simply increasing the model size.

\paragraph{Experiments on Video Datasets}
Our model is designed for anomaly detection in images, but to validate the generalization of our method, we conduct experiments on two real-world video anomaly detection datasets, i.e. UCSD-Ped2~\cite{li2013anomaly} and CUHK Avenue~\cite{lu2013abnormal}. Following the experimental setting in~\cite{park2020learning}, we resize each frame of the video into the size of $256 \times 256$, and normalize the pixel values to the range of $[-1, 1]$. $r_{h}, r_{w}, r_{c}$ and the memory bank size $N$ are empirically set to 16, 16, 1 and 2000. The model is optimized via the Adam optimizer with a fixed learning rate of $2 e^{-4}$, a weight decay of 0, momentums of $\beta_{1}=0.9$, $\beta_{2}=0.999$, and a batch size of 4 for 60 epochs on both UCSD Ped2 and CUHK Avenue. As shown in Table~\ref{tab-video}, compared with other reconstruction-based methods, our model can still achieve competitive results. It is worth noting that our model has not been tailored for video data as MemAE does, where 3D convolution is utilized to exploit the temporal information in videos. We leave such improvement as future work.

\paragraph{Structure of DAAD}
Table~\ref{daad} shows the structure of DAAD. We use the skip connection to improve the reconstruction ability of our model and furthermore utilize the block-wise memomry module to balance the reconstruction of the normality and anomaly. 

\paragraph{Structure of the Discriminator}
Table~\ref{discriminator} shows the structure of the discriminator used in DAAD+. Its architecture is designed by following the discriminator of DCGAN~\cite{radford2015unsupervised}. The dimension of the flattened output representation from the feature extractor is 100 if the size of the input image is $256 \times 256$.

\begin{table}[]
	\begin{center}
		\begin{tabular}{l|cc}
			\hline
			Method\textbackslash{}Dataset & UCSD-Ped2 & Avenue \\ \hline
			AE-Conv2D ~\cite{hasan2016learning}                     & 0.850      & 0.800    \\
			AE-Conv3D~\cite{zhao2017spatio}                     & 0.912     & 0.771  \\
			TSC~\cite{luo2017revisit}                             & 0.910      & 0.806  \\
			StackRNN~\cite{luo2017revisit}                      & 0.922     & 0.817  \\
			AbnormalGAN~\cite{ravanbakhsh2017abnormal}                   & 0.935     & -      \\
			AE~\cite{gong2019memorizing}                            & 0.917     & 0.810   \\
			MemAE~\cite{gong2019memorizing}                         & 0.941     & 0.833  \\
			AE~\cite{park2020learning}                            & 0.864     & 0.806  \\
			MNAD~\cite{park2020learning}                          & 0.902     & 0.828  \\ \hline
			AE                            & 0.901     & 0.805   \\
			Ours                          & 0.941     & 0.829  \\ \hline
		\end{tabular}
	\end{center}
	\caption{Comparison with existing methods on two video datasets (UCSD-Ped2 and CUHK Avenue) in terms of AUROC.}
	\label{tab-video}
\end{table}

\begin{table*}[]
	\begin{center}
		\footnotesize
		\begin{tabular}{l|l|l|l|l}
			\hline
			~Layer         & ~Filter & ~Channels & ~Input             & ~Output                \\ \hline
			~ConvBNReLU    & ~$3\times3, s1$    & ~$64$      & ~$input$      & ~$enc_{1-1}~( H\times W)$          \\ \hline
			~ConvBNReLU    & ~$3\times3, s1$    & ~$64$      & ~$enc_{1-1}$       & ~$enc_{1-2}~( H\times W)$          \\ \hline 
			~Maxpooling    & ~$2\times 2, s2$    & ~$64$      & ~$enc_{1-2}$    & ~$enc_{1-3}~( H/2\times W/2)$      \\ \hline
			~BW Memory1    & ~-      & ~-      & ~$enc_{1-3}$              & ~$m_{1}~( H/2\times W/2)$      \\ \hline
			~ConvBNReLU    & ~$3\times3, s1$    & ~$128$     & ~$enc_{1-3}$              & ~$enc_{2-1}~( H/2\times W/2)$     \\ \hline 
			~ConvBNReLU    & ~$3\times3, s1$    & ~$128$     & ~$enc_{2-1}$              & ~$enc_{2-2}~(H/2\times W/2)$     \\ \hline
			~Maxpooling    & ~$2\times 2, s2$    & ~$128$     & ~$enc_{2-2}$              & ~$enc_{2-3}~( H/4\times W/4)$     \\ \hline
			~BW Memory2    & ~-      & ~-     & ~$enc_{2-3}$              & ~$m_{2}~( H/4\times W/4)$     \\ \hline
			~ConvBNReLU    & ~$3\times3, s1$    & ~$256$     & ~$enc_{2-3}$             & ~$enc_{3-1}~( H/4\times W/4)$     \\ \hline 
			~ConvBNReLU    & ~$3\times3,s1$    & ~$256$     & ~$enc_{3-1}$              & ~$enc_{3-2}~( H/4\times W/4)$     \\ \hline 
			~Maxpooling    & ~$2\times 2, s2$    & ~$256$     & ~$enc_{3-2}$              & ~$enc_{3-3}~( H/8\times W/8)$     \\ \hline
			~BW Memory3    & ~-      & ~-     & ~$enc_{3-3}$              & ~$m_{3}~( H/8\times W/8)$     \\ \hline
			~ConvBNReLU    & ~$3\times3, s1$    & ~$512$     & ~$enc_{3-3}$              & ~$enc_{4-1}~( H/8\times W/8)$     \\ \hline
			~ConvBNReLU    & ~$3\times3, s1$    & ~$512$     & ~$enc_{4-1}$              & ~$enc_{4-2}~( H/8\times W/8)$     \\ \hline
			~Maxpooling    & ~$2\times 2, s2$    & ~$512$     & ~$enc_{4-2}$              & ~$enc_{4-3}~( H/16\times W/16)$   \\ \hline
			~BW Memory4    & ~-      & ~-     & ~$enc_{4-3}$              & ~$m_{4}~( H/16\times W/16)$   \\ \hline
			~ConvBNReLU    & ~$3\times3, s1$    & ~$1024$    & ~$m_{4}$              & ~$dec_{4-1}~( H/16\times W/16)$ \\ \hline 
			~ConvBNReLU    & ~$3\times3, s1$    & ~$1024$    & ~$dec_{4-1}$             & ~$dec_{4-2}~( H/16\times W/16)$ \\ \hline 
			~ConvTranspose & ~$2\times 2, s2$    & ~$512$     & ~$dec_{4-2}$             & ~$dec_{4-3}~( H/8\times W/8)$    \\ \hline
			~ConvBNReLU    & ~$3\times3, s1$    & ~$512$     & ~$[dec_{4-3}, m_{3}]$ & ~$dec_{3-1}~( H/8\times W/8)$    \\ \hline 
			~ConvBNReLU    & ~$3\times3, s1$    & ~$512$     & ~$dec_{3-1}$             & ~$dec_{3-2}~( H/8\times W/8)$    \\ \hline 
			~ConvTranspose & ~$2\times 2, s2$    & ~$256$     & ~$dec_{3-2}$             & ~$dec_{3-3}~( H/4\times W/4)$    \\ \hline
			~ConvBNReLU    & ~$3\times3, s1$    & ~$256$     & ~$[dec_{3-3}, m_{2}]$ & ~$dec_{2-1}~( H/4 \times W/4)$    \\ \hline 
			~ConvBNReLU    & ~$3\times3, s1$    & ~$256$     & ~$dec_{2-1}$             & ~$dec_{2-2}~( H/4\times W/4)$    \\ \hline 
			~ConvTranspose & ~$2\times 2, s2$    & ~$128$     & ~$dec_{2-2}$             & ~$dec_{2-3}~( H/2\times W/2)$    \\ \hline
			~ConvBNReLU    & ~$3\times3, s1$    & ~$128$     & ~$[dec_{2-3}, m_{1}]$ & ~$dec_{1-1}~( H/2\times W/2)$    \\ \hline
			~ConvBNReLU    & ~$3\times 3, s1$    & ~$128$     & ~$dec_{1-1}$             & ~$dec_{1-2}~( H/2\times W/2)$    \\ \hline
			~ConvTranspose & ~$2\times 2, s2$    & ~$64$      & ~$dec_{1-2}$             & ~$dec_{1-3}~( H\times W)$         \\ \hline
			~ConvBNReLU    & ~$3\times3, s1$    & ~$64$      & ~$dec_{1-3}$             & ~$head_{1-1}~( H\times W)$       \\ \hline
			~ConvBNReLU    & ~$3\times3, s1$    & ~$64$      & ~$head_{1-1}$           & ~$head_{1-2}~( H\times W)$       \\ \hline
			~Conv          & ~$1\times 1, s1$    & ~$3$       & ~$head_{1-2}$           & ~$output~( H\times W)$         \\ \hline
		\end{tabular}
	\end{center}
	\caption{Structure of DAAD.}
	\label{daad}
\end{table*}

\newcommand{\tabincell}[2]{\begin{tabular}{@{}#1@{}}#2\end{tabular}} 
\begin{table}[]
	\begin{center}
		\footnotesize
		\begin{tabular}{c|c|c|c|c}
			\hline
			Name                                & Layer     & Filter      & Channels & Stride          \\ \hline
			\multirow{18}{*}{\tabincell{c}{Feature\\ Extractor}} & Conv      & $4\times 4$         &$64$    & $2$         \\ \cline{2-5} 
			& LeakyReLU & \multicolumn{3}{c}{$negative\_slope=0.2$} \\ \cline{2-5} 
			& Conv      & $4\times 4$          &$128$ &   $2$                  \\ \cline{2-5} 
			& BatchNorm & \multicolumn{3}{c}{-}                   \\ \cline{2-5} 
			& LeakyReLU & \multicolumn{3}{c}{$negative\_slope=0.2$} \\ \cline{2-5} 
			& Conv      & $4\times 4$        &$256$ &     $2$                  \\ \cline{2-5} 
			& BatchNorm & \multicolumn{3}{c}{-}                   \\ \cline{2-5} 
			& LeakyReLU & \multicolumn{3}{c}{$negative\_slope=0.2$} \\ \cline{2-5} 
			& Conv      & $4\times 4$       &$512$ &      $2$                \\ \cline{2-5} 
			& BatchNorm & \multicolumn{3}{c}{-}                   \\ \cline{2-5} 
			& LeakyReLU & \multicolumn{3}{c}{$negative\_slope=0.2$} \\ \cline{2-5} 
			& Conv      & $4\times 4$         &$1024$ &    $2$                \\ \cline{2-5} 
			& BatchNorm & \multicolumn{3}{c}{-}                   \\ \cline{2-5} 
			& LeakyReLU & \multicolumn{3}{c}{$negative\_slope=0.2$} \\ \cline{2-5} 
			& Conv      & $4\times 4$      &$2048$ &      $2$                \\ \cline{2-5} 
			& BatchNorm & \multicolumn{3}{c}{-}                   \\ \cline{2-5} 
			& LeakyReLU & \multicolumn{3}{c}{$negative\_slope=0.2$} \\ \cline{2-5} 
			& Conv      & $4\times 4$         &$100$ &    $1$                \\ \hline
			\multirow{2}{*}{Classifier}         & Conv      & $3\times 3$          &$1$  & $1$                    \\ \cline{2-5} 
			& \multicolumn{4}{c}{Sigmoid}                         \\ \hline
		\end{tabular}
	\end{center}
	\caption{Structure of the discriminator used in DAAD+.}
	\label{discriminator}
\end{table}

\end{document}